\documentclass[10pt,twocolumn,letterpaper]{article}

\usepackage{cvpr}              

\usepackage{graphicx}
\usepackage{amsmath}
\usepackage{amssymb}
\usepackage{booktabs}

\usepackage{multirow}

\usepackage[accsupp]{axessibility}  

\usepackage[symbol]{footmisc}

\usepackage[pagebackref,breaklinks,colorlinks]{hyperref}

\usepackage[T1]{fontenc}
\usepackage[utf8]{inputenc}
\usepackage{amsmath,amssymb,amsfonts,mathrsfs,bm}
\usepackage{mathtools}
\usepackage{amsthm}
\usepackage{scalerel}
\usepackage{nicefrac}
\usepackage{microtype} 
\usepackage[shortlabels]{enumitem}
\usepackage{graphicx}
\usepackage{epstopdf}
\DeclareGraphicsExtensions{.eps,.png,.jpg,.pdf}

\usepackage{url}
\usepackage{colortbl}
\usepackage{booktabs}
\usepackage{multirow}
\usepackage[normalem]{ulem}
\usepackage{xparse}
\usepackage{calc}
\usepackage{etoolbox}

\makeatletter
\@ifpackageloaded{natbib}{
	\relax
}{
	\usepackage{cite}
}
\makeatother

\usepackage{array}
\newcolumntype{L}[1]{>{\raggedright\let\newline\\\arraybackslash\hspace{0pt}}m{#1}}
\newcolumntype{C}[1]{>{\centering\let\newline\\\arraybackslash\hspace{0pt}}m{#1}}
\newcolumntype{R}[1]{>{\raggedleft\let\newline\\\arraybackslash\hspace{0pt}}m{#1}}

\makeatletter
\let\MYcaption\@makecaption
\makeatother
\usepackage[font=footnotesize]{subcaption}
\makeatletter
\let\@makecaption\MYcaption
\makeatother

\usepackage{glossaries}
\makeatletter
\sfcode`\.1006

\let\oldgls\gls
\let\oldglspl\glspl

\newcommand\fussy@ifnextchar[3]{%
	\let\reserved@d=#1%
	\def\reserved@a{#2}%
	\def\reserved@b{#3}%
	\futurelet\@let@token\fussy@ifnch}
\def\fussy@ifnch{%
	\ifx\@let@token\reserved@d
		\let\reserved@c\reserved@a
	\else
		\let\reserved@c\reserved@b
	\fi
	\reserved@c}

\renewcommand{\gls}[1]{%
\oldgls{#1}\fussy@ifnextchar.{\@checkperiod}{\@}}
\renewcommand{\glspl}[1]{%
\oldglspl{#1}\fussy@ifnextchar.{\@checkperiod}{\@}}

\newcommand{\@checkperiod}[1]{%
	\ifnum\sfcode`\.=\spacefactor\else#1\fi
}

\robustify{\gls}
\robustify{\glspl}
\makeatother

\newacronym{wrt}{w.r.t.}{with respect to}
\newacronym{RHS}{R.H.S.}{right-hand side}
\newacronym{LHS}{L.H.S.}{left-hand side}
\newacronym{iid}{i.i.d.}{independent and identically distributed}

\usepackage[capitalize]{cleveref}
\crefname{equation}{}{}
\Crefname{equation}{}{}
\crefname{claim}{claim}{claims}
\crefname{step}{step}{steps}
\crefname{line}{line}{lines}
\crefname{condition}{condition}{conditions}
\crefname{dmath}{}{}
\crefname{dseries}{}{}
\crefname{dgroup}{}{}

\crefname{Problem}{Problem}{Problems}
\crefformat{Problem}{Problem~(#2#1#3)}
\crefrangeformat{Problem}{Problems~(#3#1#4) to~(#5#2#6)}

\crefname{Theorem}{Theorem}{Theorems}
\crefname{Corollary}{Corollary}{Corollaries}
\crefname{Proposition}{Proposition}{Propositions}
\crefname{Lemma}{Lemma}{Lemmas}
\crefname{Definition}{Definition}{Definitions}
\crefname{Example}{Example}{Examples}
\crefname{Assumption}{Assumption}{Assumptions}
\crefname{Remark}{Remark}{Remarks}
\crefname{Rem}{Remark}{Remarks}
\crefname{remarks}{Remarks}{Remarks}
\crefname{Appendix}{Appendix}{Appendices}
\crefname{Supplement}{Supplement}{Supplements}
\crefname{Exercise}{Exercise}{Exercises}
\crefname{Theorem_A}{Theorem}{Theorems}
\crefname{Corollary_A}{Corollary}{Corollaries}
\crefname{Proposition_A}{Proposition}{Propositions}
\crefname{Lemma_A}{Lemma}{Lemmas}
\crefname{Definition_A}{Definition}{Definitions}

\usepackage{crossreftools}

\usepackage{algorithm}
\usepackage{algpseudocode}

\ifx\loadbreqn\undefined
	\relax
\else
	\usepackage{breqn}
\fi

\makeatletter
\def\cleartheorem#1{%
    \expandafter\let\csname#1\endcsname\relax
    \expandafter\let\csname c@#1\endcsname\relax
}
\def\clearthms#1{ \@for\tname:=#1\do{\cleartheorem\tname} }
\makeatother

\ifx\renewtheorem\undefined
	\ifx\useTheoremCounter\undefined
		\newtheorem{Theorem}{Theorem}
		\newtheorem{Corollary}{Corollary}
		\newtheorem{Proposition}{Proposition}
		
	\else

	\fi

\fi

\theoremstyle{remark}

\theoremstyle{plain}

\newcommand{\qednew}{\nobreak \ifvmode \relax \else
		\ifdim\lastskip<1.5em \hskip-\lastskip
			\hskip1.5em plus0em minus0.5em \fi \nobreak
		\vrule height0.75em width0.5em depth0.25em\fi}



\newcommand{\ml}[1]{\begin{multlined}[t]#1\end{multlined}}

\NewDocumentCommand{\movedownsub}{e{^_}}{%
	\IfNoValueTF{#1}{%
		\IfNoValueF{#2}{^{}}
	}{%
		^{#1}
	}%
	\IfNoValueF{#2}{_{#2}}
}

\let\latexchi\chi
\RenewDocumentCommand{\chi}{}{\latexchi\movedownsub}

\newcommand{\bbR}{\mathbb{R}}

\DeclareSymbolFont{bsfletters}{OT1}{cmss}{bx}{n}
\DeclareSymbolFont{ssfletters}{OT1}{cmss}{m}{n}
\DeclareMathSymbol{\bsfGamma}{0}{bsfletters}{'000}
\DeclareMathSymbol{\ssfGamma}{0}{ssfletters}{'000}
\DeclareMathSymbol{\bsfDelta}{0}{bsfletters}{'001}
\DeclareMathSymbol{\ssfDelta}{0}{ssfletters}{'001}
\DeclareMathSymbol{\bsfTheta}{0}{bsfletters}{'002}
\DeclareMathSymbol{\ssfTheta}{0}{ssfletters}{'002}
\DeclareMathSymbol{\bsfLambda}{0}{bsfletters}{'003}
\DeclareMathSymbol{\ssfLambda}{0}{ssfletters}{'003}
\DeclareMathSymbol{\bsfXi}{0}{bsfletters}{'004}
\DeclareMathSymbol{\ssfXi}{0}{ssfletters}{'004}
\DeclareMathSymbol{\bsfPi}{0}{bsfletters}{'005}
\DeclareMathSymbol{\ssfPi}{0}{ssfletters}{'005}
\DeclareMathSymbol{\bsfSigma}{0}{bsfletters}{'006}
\DeclareMathSymbol{\ssfSigma}{0}{ssfletters}{'006}
\DeclareMathSymbol{\bsfUpsilon}{0}{bsfletters}{'007}
\DeclareMathSymbol{\ssfUpsilon}{0}{ssfletters}{'007}
\DeclareMathSymbol{\bsfPhi}{0}{bsfletters}{'010}
\DeclareMathSymbol{\ssfPhi}{0}{ssfletters}{'010}
\DeclareMathSymbol{\bsfPsi}{0}{bsfletters}{'011}
\DeclareMathSymbol{\ssfPsi}{0}{ssfletters}{'011}
\DeclareMathSymbol{\bsfOmega}{0}{bsfletters}{'012}
\DeclareMathSymbol{\ssfOmega}{0}{ssfletters}{'012}

\makeatletter
\newcommand*\rel@kern[1]{\kern#1\dimexpr\macc@kerna}
\newcommand*\widebar[1]{%
  \begingroup
  \def\mathaccent##1##2{%
    \rel@kern{0.8}%
    \overline{\rel@kern{-0.8}\macc@nucleus\rel@kern{0.2}}%
    \rel@kern{-0.2}%
  }%
  \macc@depth\@ne
  \let\math@bgroup\@empty \let\math@egroup\macc@set@skewchar
  \mathsurround\z@ \frozen@everymath{\mathgroup\macc@group\relax}%
  \macc@set@skewchar\relax
  \let\mathaccentV\macc@nested@a
  \macc@nested@a\relax111{#1}%
  \endgroup
}
\makeatother

\DeclareMathOperator*{\concat}{\scalerel*{\parallel}{\sum}}

\newcommand{\ifbcdot}[1]{\ifblank{#1}{\cdot}{#1}}

\DeclarePairedDelimiterX\abs[1]{\lvert}{\rvert}{\ifbcdot{#1}}
\DeclarePairedDelimiterX\parens[1]{(}{)}{\ifbcdot{#1}}
\DeclarePairedDelimiterX\brk[1]{[}{]}{\ifbcdot{#1}}
\DeclarePairedDelimiterX\braces[1]{\{}{\}}{\ifbcdot{#1}}
\DeclarePairedDelimiterX\angles[1]{\langle}{\rangle}{\ifblank{#1}{\cdot,\cdot}{#1}}
\DeclarePairedDelimiterX\ip[2]{\langle}{\rangle}{\ifbcdot{#1},\ifbcdot{#2}}
\DeclarePairedDelimiterX\norm[1]{\lVert}{\rVert}{\ifbcdot{#1}}
\DeclarePairedDelimiterX\ceil[1]{\lceil}{\rceil}{\ifbcdot{#1}}
\DeclarePairedDelimiterX\floor[1]{\lfloor}{\rfloor}{\ifbcdot{#1}}

\DeclareFontFamily{U}{matha}{\hyphenchar\font45}
\DeclareFontShape{U}{matha}{m}{n}{
      <5> <6> <7> <8> <9> <10> gen * matha
      <10.95> matha10 <12> <14.4> <17.28> <20.74> <24.88> matha12
      }{}
\DeclareSymbolFont{matha}{U}{matha}{m}{n}
\DeclareFontSubstitution{U}{matha}{m}{n}

\DeclareFontFamily{U}{mathx}{\hyphenchar\font45}
\DeclareFontShape{U}{mathx}{m}{n}{
      <5> <6> <7> <8> <9> <10>
      <10.95> <12> <14.4> <17.28> <20.74> <24.88>
      mathx10
      }{}
\DeclareSymbolFont{mathx}{U}{mathx}{m}{n}
\DeclareFontSubstitution{U}{mathx}{m}{n}

\DeclareMathDelimiter{\vvvert}{0}{matha}{"7E}{mathx}{"17}
\DeclarePairedDelimiterX\vertiii[1]{\vvvert}{\vvvert}{\ifbcdot{#1}}

\DeclarePairedDelimiterXPP\trace[1]{\operatorname{Tr}}{(}{)}{}{\ifbcdot{#1}} 
\DeclarePairedDelimiterXPP\col[1]{\operatorname{col}}{\{}{\}}{}{\ifbcdot{#1}} 
\DeclarePairedDelimiterXPP\row[1]{\operatorname{row}}{\{}{\}}{}{\ifbcdot{#1}} 
\DeclarePairedDelimiterXPP\erf[1]{\operatorname{erf}}{(}{)}{}{\ifbcdot{#1}}
\DeclarePairedDelimiterXPP\erfc[1]{\operatorname{erfc}}{(}{)}{}{\ifbcdot{#1}}
\DeclarePairedDelimiterXPP\KLD[2]{D}{(}{)}{}{\ifbcdot{#1}\, \delimsize\|\, \ifbcdot{#2}} 
\DeclarePairedDelimiterXPP\op[2]{\operatorname{#1}}{(}{)}{}{#2} 

\newcommand{\T}{^{\intercal}}

\DeclarePairedDelimiterXPP\indicate[1]{{\bf 1}}{\{}{\}}{}{\ifbcdot{#1}}

\providecommand\given{}

\DeclarePairedDelimiterX\Set[2]\{\}{%
\renewcommand\given{\SetSymbol[\delimsize]{#1}}
#2
}
\DeclarePairedDelimiterX\Setc[1]\{\}{%
\renewcommand\given{\SetSymbol{:}}
#1
}

\NewDocumentCommand\set{s o m}{%
	\IfBooleanTF#1%
	{\IfValueTF{#2}{\Set*{#2}{#3}}{\Setc*{#3}}}%
	{\IfValueTF{#2}{\Set{#2}{#3}}{\Setc{#3}}}%
}

\NewDocumentCommand{\evalat}{ s O{\big} m e{_^} }{%
\IfBooleanTF{#1}%
{\left. #3 \right|}{#3#2|}%
\IfValueT{#4}{_{#4}}%
\IfValueT{#5}{^{#5}}%
}

\providecommand\given{}
\DeclarePairedDelimiterXPP\cprob[1]{}(){}{
\renewcommand\given{\nonscript\,\delimsize\vert\allowbreak\nonscript\,\mathopen{}}%
#1%
}
\DeclarePairedDelimiterXPP\cexp[1]{}[]{}{
\renewcommand\given{\nonscript\,\delimsize\vert\allowbreak\nonscript\,\mathopen{}}%
#1%
}

\DeclareDocumentCommand \P { s e{_^} d() g } {%
	\mathbb{P}%
	\IfBooleanTF{#1}%
		{
			\IfValueT{#2}{_{#2}}%
			\IfValueT{#3}{^{#3}}%
			\IfValueTF{#5}{\cprob{#4 \given #5}}{\IfValueT{#4}{\cprob{#4}}}%
		}%
		{
			\IfValueT{#2}{_{#2}}%
			\IfValueT{#3}{^{#3}}%
			\IfValueTF{#5}{\cprob*{#4 \given #5}}{\IfValueT{#4}{\cprob*{#4}}}%
		}%
}

\DeclareDocumentCommand \E { s e{_^} o g } {%
	\mathbb{E}%
	\IfBooleanTF{#1}%
		{
			\IfValueT{#2}{_{#2}}%
			\IfValueT{#3}{^{#3}}%
			\IfValueTF{#5}{\cexp{#4 \given #5}}{\IfValueT{#4}{\cexp{#4}}}%
		}%
		{
			\IfValueT{#2}{_{#2}}%
			\IfValueT{#3}{^{#3}}%
			\IfValueTF{#5}{\cexp*{#4 \given #5}}{\IfValueT{#4}{\cexp*{#4}}}%
		}%
}

\ExplSyntaxOn
\NewDocumentCommand \dist {m o o} {%
\mathrm{#1}\left(%
	\IfValueT{#3}{%
		\tl_if_blank:nTF{ #3 }{\cdot\, \middle|\, }{#3\, \middle|\, }%
	}
	\IfValueT{#2}{#2}%
\right)%
}
\ExplSyntaxOff

\NewDocumentCommand {\cbrace} {t+ D[]{black} D(){\widthof{#5}} m m } {%
	\begingroup%
		\color{#2}
		\IfBooleanTF{#1}{%
			\overbrace{#4}^%
		}{
			\underbrace{#4}_%
		}%
		{\parbox[c]{#3}{\centering\footnotesize{#5}}}%
	\endgroup%
}

\let\oldforall\forall
\renewcommand{\forall}{\oldforall \, }

\let\oldexist\exists
\renewcommand{\exists}{\oldexist \, }

\graphicspath{{./Figures/}{./figures/}}
\DeclareDocumentCommand{\includeCroppedPdf}{ o O{./Figures/} m }{
	\IfFileExists{#2#3-crop.pdf}{}{%
		\immediate\write18{pdfcrop #2#3.pdf #2#3-crop.pdf}}%
	\includegraphics[#1]{#2#3-crop.pdf}
}

\makeatletter
\newcommand*{\addFileDependency}[1]{
  \typeout{(#1)}
  \@addtofilelist{#1}
  \IfFileExists{#1}{}{\typeout{No file #1.}}
}
\makeatother

\definecolor{gray90}{gray}{0.9}

\ifx\nohighlights\undefined

	\newcommand{\msout}[1]{\text{\color{green} \sout{\ensuremath{#1}}}}
	
	\newcommand{\del}[1]{{\color{green}\ifmmode \msout{#1}\else\sout{#1}\fi}}
\else

	\newcommand{\msout}[1]{#1}
	\newcommand{\del}[1]{#1}
\fi

\newcommand{\hhide}[1]{}

\ifx\diagnoselabel\undefined
	\relax
\else
	\makeatletter
	\def\@testdef #1#2#3{%
		\def\reserved@a{#3}\expandafter \ifx \csname #1@#2\endcsname
			\reserved@a  \else
			\typeout{^^Jlabel #2 changed:^^J%
				\meaning\reserved@a^^J%
				\expandafter\meaning\csname #1@#2\endcsname^^J}%
			\@tempswatrue \fi}
	\makeatother
\fi

\DeclareFontFamily{U}{MnSymbolC}{}
\DeclareSymbolFont{MnSyC}{U}{MnSymbolC}{m}{n}
\DeclareMathSymbol{\diamondplus}{\mathbin}{MnSyC}{"7C}
\DeclareMathSymbol{\diamonddot}{\mathbin}{MnSyC}{"7E}
\DeclareFontShape{U}{MnSymbolC}{m}{n}{
    <-6>  MnSymbolC5
   <6-7>  MnSymbolC6
   <7-8>  MnSymbolC7
   <8-9>  MnSymbolC8
   <9-10> MnSymbolC9
  <10-12> MnSymbolC10
  <12->   MnSymbolC12}{}

\DeclareMathOperator{\arctanh}{arctanh}

\def\cvprPaperID{10726} 
\def\confName{CVPR}
\def\confYear{2023}

\begin{document}

\title{HypLiLoc: Towards Effective LiDAR Pose Regression with Hyperbolic Fusion}

\author{
Sijie Wang\textsuperscript{\rm 1}\textsuperscript{*}\quad
Qiyu Kang\textsuperscript{\rm 1}\textsuperscript{*}\quad
Rui She\textsuperscript{\rm 1}\quad
Wei Wang\textsuperscript{\rm 2}\quad
Kai Zhao\textsuperscript{\rm 1}\quad
Yang Song\textsuperscript{\rm 3}\quad
Wee Peng Tay\textsuperscript{\rm 1}\and
\textsuperscript{\rm 1}Nanyang Technological University\quad
\textsuperscript{\rm 2}University of Oxford\quad
\textsuperscript{\rm 3}C3.AI\\
{\tt\small\{wang1679;qiyu.kang;rui.she;kai.zhao;wptay\}@ntu.edu.sg}\\
{\tt\small peter.wang1221@gmail.com}\quad
{\tt\small yang.song@c3.ai}
}

    

\maketitle

\begin{abstract}
LiDAR relocalization plays a crucial role in many fields, including robotics, autonomous driving, and computer vision. LiDAR-based retrieval from a database typically incurs high computation storage costs and can lead to globally inaccurate pose estimations if the database is too sparse. On the other hand, pose regression methods take images or point clouds as inputs and directly regress global poses in an end-to-end manner. They do not perform database matching and are more computationally efficient than retrieval techniques. We propose HypLiLoc, a new model for LiDAR pose regression. We use two branched backbones to extract 3D features and 2D projection features, respectively. We consider multi-modal feature fusion in both Euclidean and hyperbolic spaces to obtain more effective feature representations. Experimental results indicate that HypLiLoc achieves state-of-the-art performance in both outdoor and indoor datasets. We also conduct extensive ablation studies on the framework design, which demonstrate the effectiveness of multi-modal feature extraction and multi-space embedding. Our code is released at: \url{https://github.com/sijieaaa/HypLiLoc}
\end{abstract}

\section{Introduction}
\footnotetext[1]{These authors contribute equally. This paper has been accepted by CVPR 2023.}
Visual relocalization aims at estimating the 6-degree of freedom (DoF) pose of an agent using perception sensors, such as LiDARs and cameras. It plays a crucial role in many fields that include robot navigation \cite{navigation}, autonomous driving \cite{l3net}, and scene recognition \cite{topological}. 
Image-based relocalization methods have achieved good performance in various applications \cite{wang2020atloc, henriques2018mapnet, ms-transformer}. However, images taken from cameras can only capture RGB color information and are easily influenced by environmental conditions, including low illumination and light reflections. By contrast, LiDARs, which cast active beams to estimate the depth of surrounding objects, are more robust against those changes.

In recent years, the LiDAR has become an important sensor in smart robots, autonomous vehicles, and mobile devices. LiDAR-based relocalization, which is a basic and important module impacting other perception tasks, has attracted more attention \cite{zhang2017low,wang2021f,koide2021voxelized,dellenbach2022ct, nubert2021self,li2021self}. One of the classical approaches, LiDAR odometry, estimates the relative poses among successive LiDAR frames to obtain locally accurate pose estimation. However, errors accumulate over the trajectory, resulting in unsatisfactory global pose estimation. To compensate for the error, LiDAR odometry is usually treated as a component in a complete simultaneous localization and mapping system (SLAM), where the global pose estimated by a global positioning method or detected loop closure is used to correct the accumulated error in the LiDAR odometry \cite{shan2020lio,xu2021fast}.

LiDAR-based retrieval is also used for relocalization \cite{uy2018pointnetvlad}. It first constructs a database of LiDAR features learned from all candidate LiDAR frames. During inference, given a query LiDAR scan, the similarities between the query feature and all features stored in the database are computed so that the top-matched poses can be obtained. Although this approach provides accurate global pose estimation, it inherently suffers from high computation cost and storage burden \cite{wang2021pointloc}. Therefore, it is more appropriate for offline scenarios rather than for real-time mobile applications. 

Pose regression is favored as a relocalization method due to its lower computation and storage cost during inference. The pose regression network is still trained on a database containing LiDAR frames in an end-to-end manner to obtain a regression model.
During inference, taking the LiDAR scan as input, the pose regression network directly regresses the global pose without any pre-constructed candidate database or map. It can mitigate the high computation and storage burden that occurs in the LiDAR-based retrieval methods. As a result, pose regression can be operated in real-time to satisfy various relocalization requirements in robotics, unmanned aerial vehicles (UAVs), mobile relocalization APPs, autonomous vehicles, and SLAM systems.

In this paper, we propose a relocalization method called HypLiLoc, which is a pose regression network with LiDAR data as input. HypLiLoc uses a parallel feature extraction design, in which 3D features and 2D spherical projection features are obtained in two backbone branches simultaneously. The paper \cite{montanaro2022hyp-pointcloud} leverages hyperbolic embeddings for 3D point clouds that can be viewed as hierarchical compositions of small parts. We thus follow this motivation to design our pipeline with hyperbolic learning. 
Specifically, we conduct feature fusion in both Euclidean and hyperbolic spaces to enhance the information representation and to achieve more effective multi-modal feature interaction. We test HypLiLoc in both outdoor and indoor datasets. Experiments indicate that HypLiLoc surpasses current approaches and achieves state-of-the-art (SOTA) performance.

Our main contributions are summarized as follows:
\begin{enumerate}
\item We propose a novel LiDAR-based pose regression network HypLiLoc. It has one backbone that learns 3D features directly from the 3D point cloud and another backbone that learns features from a 2D projection of the point cloud onto a spherical surface.
To achieve effective multi-modal feature interaction, the features are embedded in both Euclidean and hyperbolic spaces using multi-space learning. An attention mechanism is then used to fuse the features from different spaces together.
\item We test our network in both outdoor and indoor datasets, where it outperforms current LiDAR pose regression counterparts and achieves SOTA performance. We also conduct extensive ablation studies on the effectiveness of each design component.
\end{enumerate}

\section{Related Work}
In this section, we shall introduce more relocalization works, including LiDAR odometry, point cloud retrieval, and pose regression. Besides, we shall also provide more details of hyperbolic learning that are related to our method design.

\subsection{LiDAR Odometry}
LiDAR odometry methods address the relocalization problem under local views. Given several close or nearby LiDAR scans, they estimate the relative poses among them. 
The iterative closest point (ICP) method \cite{besl1992method} solves the relative pose by iteratively searching correspondence between source points and target points and optimizing the least square error. Besides, another popular method is LOAM \cite{zhang2017low}, which classifies the keypoints into edges and planes and uses KD-trees to search the neighborhood of the keypoints. DCP \cite{Wang_2019_ICCV} uses PointNet \cite{qi2017pointnet} and DGCNN \cite{wang2019dynamic} as backbones to extract point cloud features, and then it predicts the pose using the Transformer module.

\subsection{Point Cloud Retrieval}
Point cloud retrieval approaches treat the relocalization task as the place recognition problem \cite{uy2018pointnetvlad}. The core of these approaches is query-database matching. A database needs to be constructed to store the features of all candidate LiDAR scans with corresponding poses. In the inference process, for a given query LiDAR scan, its feature is extracted by the neural network, and then the query-database feature matching is performed for every possible pair. The final pose estimation is obtained from the top-matched pairs.

\subsection{Pose Regression}
Given the query sensor data, the pose regression models directly outputs the pose using the trained neural network. They do not depend on the query-database matching procedure, which speeds up the inference stage significantly when compared to point cloud retrieval methods. These models are still trained on a database of training samples that include sensor scans and the ground truth sensor poses. However, during the inference stage, the database is no longer required, in contrast to retrieval methods.

PoseNet \cite{kendall2015posenet} proposes simultaneous learning for location and orientation by integrating balance parameters. MapNet\cite{henriques2018mapnet} uses visual odometry as the post-processing technique to optimize the regressed poses. AD-PoseNet \cite{ad-posenet} leverages semantic masks to drop out the dynamic area in the image. AtLoc \cite{wang2020atloc} introduces global attention to guide the network to learn better representations. MS-Transformer \cite{ms-transformer} focuses on simultaneous pose regression for multiple scenes using a single network. RobustLoc\cite{wang2022robustloc} leverages multi-view images for robust camera pose regression.

A LiDAR actively casts beams to estimate the sparse depth of its surrounding environment. Since LiDARs are less likely to be influenced by illumination changes than cameras, they have become core sensors in many applications. PointLoc \cite{wang2021pointloc} uses LiDAR point clouds to achieve pose regression. Its model consists of PoinNet++ \cite{qi2017pointnet++} followed by self-attention modules to generate point cloud features. The paper \cite{memory-aware} studies the memory-friendly pose regression learning scheme and proposes four LiDAR pose regression models.

\begin{figure*}
\begin{center}
\includegraphics[width=0.95\textwidth]{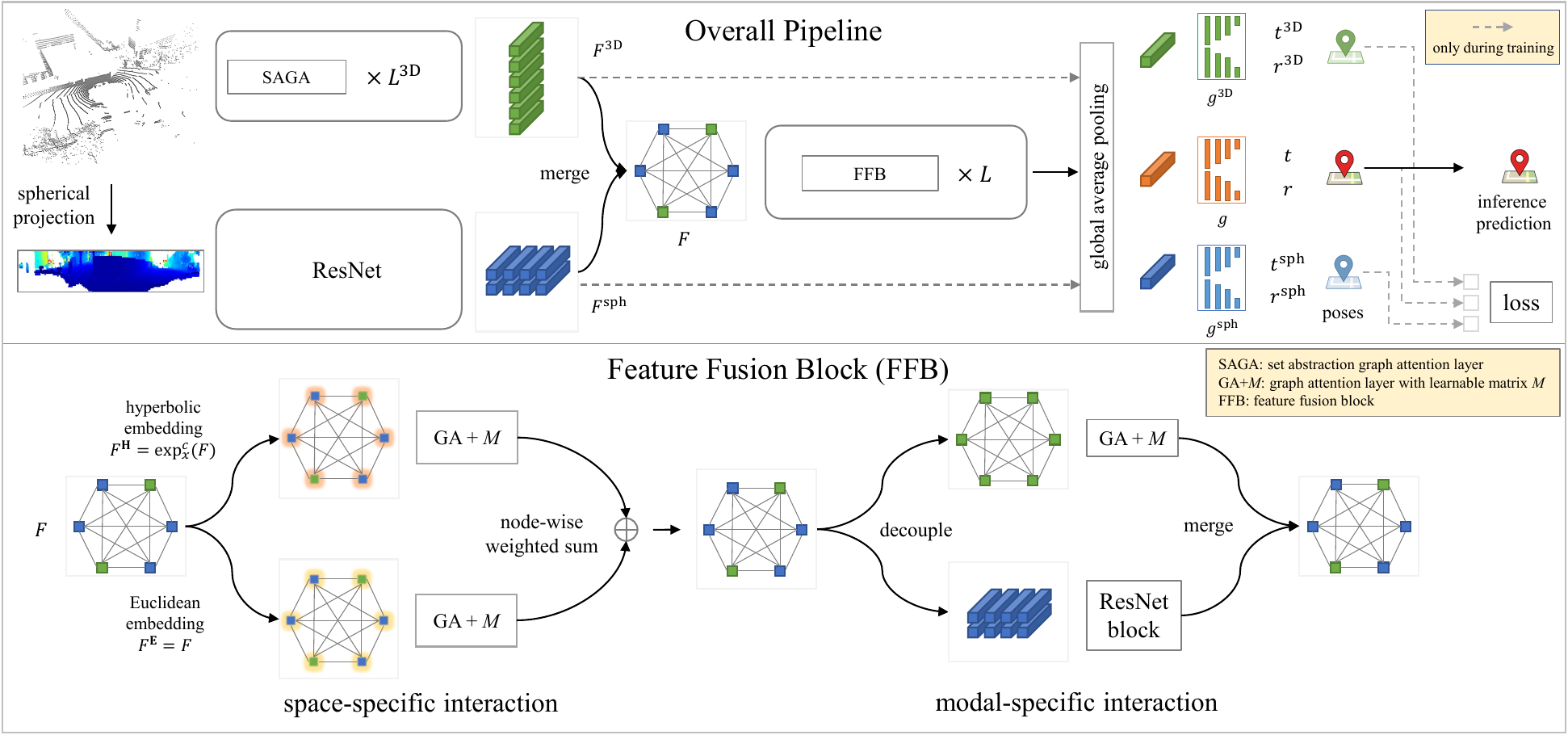}
\end{center}
\caption{The overall architecture of our proposed HypLiLoc. We use two backbone branches to perform feature extraction. In the 3D backbone, we consider both local set abstraction and global attention aggregation. In the feature fusion block, the extracted multi-modal features are embedded into both Euclidean and hyperbolic spaces to achieve space-specific interaction. The fusion features are then decoupled to their own modality to perform modal-specific interaction. The final training loss is applied on both the 3D/projection level and the final fusion level.}
\label{fig:architecture}
\end{figure*}

\subsection{Hyperbolic Learning}
Hyperbolic embedding for features has been proposed for datasets that have some underlying tree structure \cite{sarkar2011low}. The paper \cite{hnn} derives hyperbolic versions of several deep learning tools, including multinomial logistic regression, feed-forward networks, and recurrent networks. In the field of natural language processing (NLP), \cite{nickel2017nlp1} and \cite{nickel2018nlp2} introduce feature embeddings with hyperbolic models.  In the field of deep graph learning, HGCN \cite{hgcn} considers hyperbolic node embeddings in Graph Convolutional Neural Networks (GCNs). GIL \cite{gil} proposes to use weighted embedding features in both Euclidean and hyperbolic spaces. In the computer vision community, \cite{hyp_img} uses pair-wise cross-entropy loss with hyperbolic distances to train the vision transformer \cite{vit}, and \cite{hyp_seg} considers hyperbolic embeddings in the semantic segmentation task. More recently, hyperbolic embeddings have also been studied for the 3D point cloud \cite{montanaro2022hyp-pointcloud}, where the 3D point cloud is treated as nature compositions of small parts that follow the hierarchical architecture. This motivates us to introduce hyperbolic embeddings in our pipeline for better feature representations.

\section{Proposed Model}
In this section, we provide a detailed description of our proposed approach. We first summarize the HypLiLoc pipeline as follows.
\begin{enumerate}
\item Given a LiDAR point cloud scan, in addition to the traditional backbone of extracting 3D features from the point cloud, we additionally project the 3D points into a sphere to generate a 2D projection image. These two types of features are extracted by separate backbones.
\item We merge the two modal features together as the fusion features. The fusion features are then embedded in both Euclidean and hyperbolic spaces to achieve more effective representations.
\item After features interact in different spaces and modalities, the global feature vector is obtained by applying the global average pooling operation on the fusion features. The final pose prediction is generated using the global feature vector with the pose regression head.
\end{enumerate}

\subsection{Modal-Specific Backbones}\label{ssec:backbone}
\begin{figure}[!htb]
\begin{center}
\includegraphics[width=0.4\textwidth]{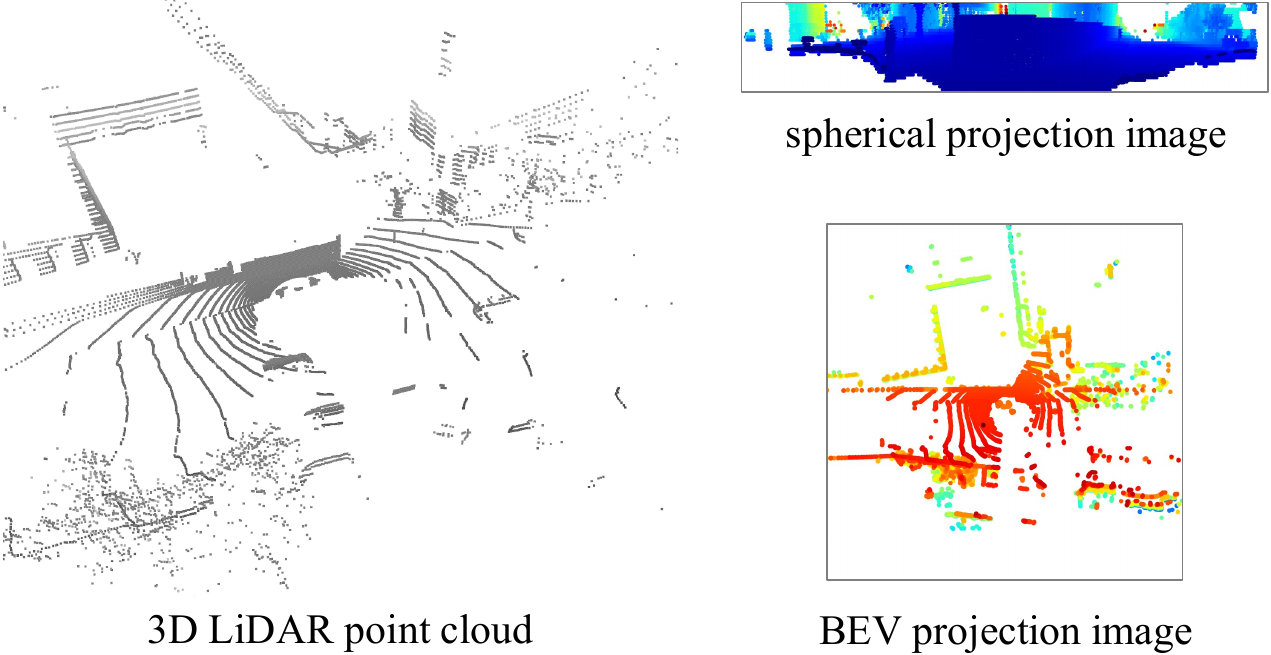}
\end{center}
\caption{Visualization of the spherical and BEV projection methods. 
}
\label{fig:sphection}
\end{figure}

\textbf{Projection Feature Extraction.}
Multi-modal feature extraction has shown promising performance in various tasks \cite{akbari2021vatt, yan20222dpass, xu2021rpvnet, clip}. The point cloud generated by LiDARs is convertible into multiple modalities by projecting 3D points into specific 2D spaces. Each projection provides us with a different way to define the neighbors of a point so that the point can aggregate feature representations from different definitions of its ``neighborhood''. To this end, we consider two typical projection methods, including the spherical projection and the bird's-eye view (BEV) projection. For the currently most commonly used multi-line spinning LiDAR, we visualize the point cloud projection in \cref{fig:sphection}.

Suppose we have as the input a set of $N$ LiDAR points $\set*{p_i}_{i=1}^N$, each represented by 3D Cartesian coordinates $p_i=\parens{x_i,y_i,z_i}\in\mathbb{R}^3$. For the spherical projection, we first convert the Cartesian coordinates into polar coordinates as: 
\begin{align}
\phi_i & = \mathrm{arctan}\parens*{ \frac{z_i}{\sqrt{x_i^2+y_i^2}} },\\
\theta_i & = \mathrm{arctan}\parens*{ \frac{y_i}{x_i}}, \\
r_i & = \sqrt{x_i^2+y_i^2+z_i^2}.
\end{align}
Each point projected on the sphere is then denoted as $p_i^{\mathrm{sph}}=\parens{\phi_i,\theta_i} \in\mathbb{R}^2$. Projecting all the $N$ points, an image $I^{\mathrm{sph}} \in \mathbb{R}^{H \times W}$ is obtained, in which each pixel contains the radius value:
\begin{align}\label{Isph}
I^{\mathrm{sph}}\parens*{ \left\lfloor\frac{\phi_i H}{2\pi}\right\rfloor, \left\lfloor\frac{\theta_i W}{2\pi}\right\rfloor } = r_i.
\end{align}
On the other hand, the BEV projects each point $p_i$ as $(x_i,y_i)$ and generates the image $I^{\mathrm{BEV}} \in \mathbb{R}^{H^{'} \times W^{'}}$, in which each pixel contains the height value:
\begin{align}\label{IBEV}
I^{\mathrm{BEV}}\parens*{ \left\lfloor\frac{y_i H^{'}}{2y_{\mathrm{max}}} \right\rfloor, \left\lfloor\frac{x_i W^{'}}{2x_{\mathrm{max}}} \right\rfloor} & = z_i,
\end{align}
where $x_{\mathrm{max}}$ and $y_{\mathrm{max}}$ are the maximum LiDAR range in the $x$ and $y$ directions, respectively.

We test the two projection counterparts in \cref{subsec:analysis}, where the spherical projection performs better than the BEV. The reason could be that LiDARs operate with the spinning mechanism, which is better modeled by the spherical projection. By contrast, the BEV projection loses information as some points are stacked on the same pixel and is thus not a bijective mapping.

In the following discussion, we use the spherical projection strategy in our pipeline. We treat the spherical projection points as the image modality input, while the 3D features are extracted directly from the point cloud in a separate backbone that we will introduce later. Following the golden rule for 2D image processing, we use ResNet \cite{resnet} as the backbone for the image modality. Denoting the ResNet backbone as $f^{\mathrm{sph}}\parens{}$, the final spherical projection features $F^{\mathrm{sph}} \in \bbR^{H^{\mathrm{sph}} \times W^{\mathrm{sph}} \times C}$ are obtained as:
\begin{align}\label{Fsph}
F^{\mathrm{sph}} = f^{\mathrm{sph}}\parens*{I^{\mathrm{sph}}}.
\end{align}

\begin{figure}[!htb]
\begin{center}
\includegraphics[width=0.45\textwidth]{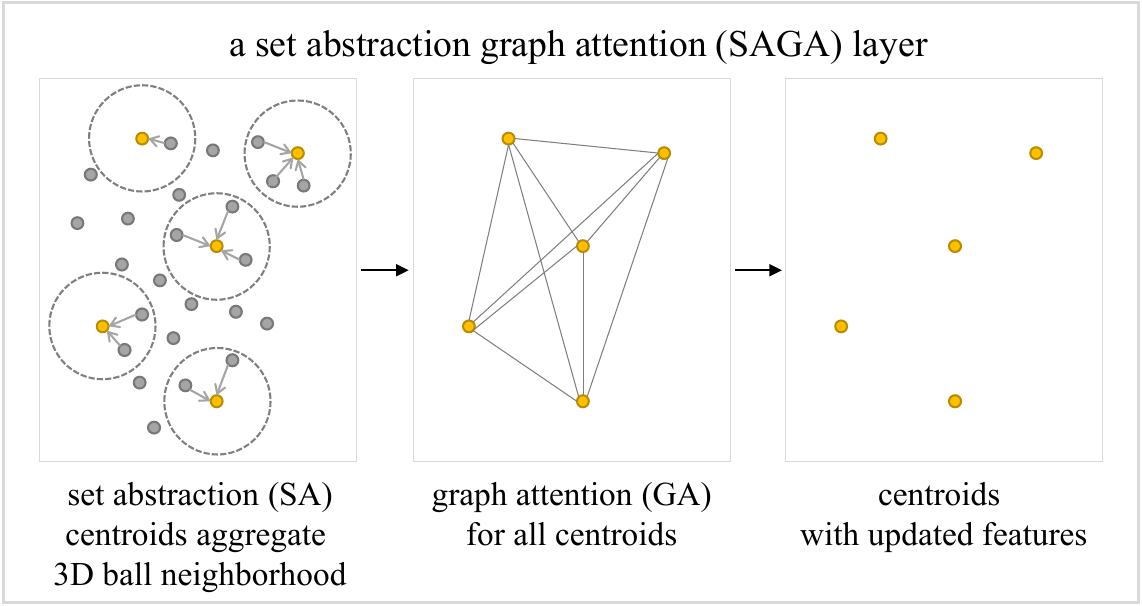}
\end{center}
\caption{The SAGA layer consists of a SA layer and a GA layer.}
\label{fig:graph_attetion}
\end{figure}

\textbf{3D Feature Extraction.}
Effective 3D point feature extraction is critical in the model design. PointNet++ \cite{qi2017pointnet++} has shown promising performance in various tasks \cite{Wang_2019_ICCV,wang2021pointloc}. In our pipeline, we use PointNet++ as the backbone branch for 3D feature extraction. 


PointNet++ only considers the neighboring information within a determined range, i.e., in the set abstraction (SA) layer, each centroid uses the maximum neighboring feature value as its updated feature as shown in \cref{fig:graph_attetion}. 
In pose regression, the estimation accuracy of the pose benefits from an effective global representation. Thus to enable PointNet++ to additionally aggregate more global information, we introduce an additional graph attention (GA) layer after each SA layer to build the set abstraction graph attention (SAGA) layer as shown in \cref{fig:graph_attetion}.
In this GA layer, we construct a complete graph whose node set contains all centroids from the SA layer.
We denote the output of the SA layer as $P \in \mathbb{R}^{N^{\mathrm{SA}} \times C^{\mathrm{SA}}}$ with $N^{\mathrm{SA}}$ centroids, each with a $C^{\mathrm{SA}}$-dimensional feature vector. 
We first use a Fully-Connected (FC) layer to generate the multi-head features: 
\begin{align}\label{eq:fc}
P_{k}^{\mathrm{FC}}=P W_{k} + b_{k},
\end{align}
where $W_{k}$ and $b_{k}$ are a linear operation and an additive bias, respectively. These are learnable parameters of the $k$-th head. Then the attention weight matrix $A_k \in \mathbb{R}^{N^{\mathrm{SA}} \times N^{\mathrm{SA}}}$ can be obtained by computing the dot product among all the neighboring nodes: 
\begin{align}\label{Ak}
A_{k}= \mathrm{Softmax}\parens*{P_k^{\mathrm{FC}}  {P_k^{\mathrm{FC}}}\T},
\end{align}
where $\mathrm{Softmax}\parens*{}$ denotes the row-wise softmax function. The output features of the GA layer $P^{\mathrm{GA}} \in \bbR^{N^{\mathrm{GA}} \times C^{\mathrm{GA}}}$ are generated by concatenating the weighted features from all heads as:
\begin{align}
P^{\mathrm{GA}}= \concat_k A_{k} P_k^{\mathrm{FC}},
\end{align}
where $\concat$ denotes the concatenation operation. We then stack $L^{\mathrm{3D}}$  such SAGA layers to build the 3D features extraction backbone. We denote the final output features as $F^{\mathrm{3D}} \in \bbR^{N^{\mathrm{3D}} \times C}$  shown in \cref{fig:architecture}. 
 


\subsection{Hyperbolic Feature Learning}\label{subsec:hyperbolic feature learning}

In this subsection, we first state the motivation for such hyperbolic feature learning. We then introduce the hyperbolic embedding operators that will be used in \cref{sect:FFB} to fuse features extracted from the 3D LiDAR point cloud and spherical LiDAR projection in \cref{ssec:backbone}.

\textbf{Motivation.}
After feature extraction using the two backbone branches, we need an effective fusion strategy to consider both point features and projection features. Embedding in a hyperbolic space has recently gained increased interest and shown promising performance in various fields \cite{gil,hgcn,hnn,hyp_img,hyp_seg}. The paper \cite{montanaro2022hyp-pointcloud} argues that 3D point cloud objects possess inherent hierarchies due to their nature as compositions of small parts, which can be embedded in the hyperbolic space. Following this motivation, we consider leveraging the hyperbolic embedding method in our pipeline, such that features can be equipped with more various representations that come from different embedding spaces. Our ablation study (cf.\ \cref{tab:ablation_study_modules} of \cref{subsec:analysis}) also indicates that hyperbolic embedding can lead to improvements in the pose estimation accuracy.


\textbf{Hyperbolic Embedding.} Different from the common Euclidean space, the hyperbolic space is equipped with constant negative curvature and has a different metric rather than the Euclidean $\ell_2$ norm $\norm{\cdot}$. 
This special metric renders a ball in hyperbolic embedding space to have exponentially increased volume with respect to its radius rather than polynomially as in Euclidean spaces. 

Similar to \cite{hyp_img}, we use the $n$-dimensional \textit{Poincar\'e ball} $(\mathbb{D}_c^n, g^{\mathbb{D}})$ for our hyperbolic embedding with the parameter $c$ indicating constant negative curvature $-c^2$. More specifically, $g^{\mathbb{D}}$ is the Riemannian metric, and $\mathbb{D}_c^n$ is defined as:
\begin{align}
\mathbb{D}_c^n =\set*{x \in \mathbb{R}^n \given c \norm{x}^2<1, c\geqslant0 },
\end{align}
where the distance $d$ between two points $x$ and $y$ on $\mathbb{D}_c^n$ is defined as:
\begin{align}
d(x,y) =\frac{2}{\sqrt{c}} \arctanh(\sqrt{c} \norm{-x\oplus_c y}), 
\end{align}
where $\oplus_c$ is the \textit{Mobius addition} defined as follows:
\begin{align}
x\oplus_c y = \frac{(1 + 2c\langle x,y \rangle + c \norm{y}^2)x + (1-c \norm{x}^2)y}{1+2c\langle x,y \rangle + c^2    \norm{x}^2 \norm{y}  ^2}.
\end{align}
To support features transferring from Euclidean spaces to hyperbolic spaces, the differentiable bijective operator named \emph{exponential map} is induced.
For a fixed base point $x\in \mathbb{D}^n_c$, where the tangent space at $x$ is a Euclidean space, the exponential map $\mathrm{exp}^c_x:\mathbb{R}^n \rightarrow \mathbb{D}^n_c$ establishes the connection between the tangent Euclidean space and the hyperbolic space at $x$ as:
\begin{align}
\mathrm{exp}^c_x(v) = x \oplus_c \parens*{\mathrm{tanh} \parens*{\sqrt{c} \frac{\lambda^c_x \norm{ v }}{2}}  \frac{v}{\sqrt{c} \norm{v} } },
\end{align}
where $\lambda^c_x$ is the \textit{conformal factor}.
In our pipeline, we assume the input features are in this tangent space and would like to embed them in the hyperbolic space $\mathbb{D}^n_c$.



\subsection{Feature Fusion Block}\label{sect:FFB}
Based on the above-mentioned motivation, we propose the feature fusion block (FFB) to achieve effective feature interaction. Each FFB conducts both space-specific interaction and modal-specific interaction alternatively, which is similar to the commonly used cross-self-attention operation. We stack $L$ such FFBs in our pipeline.

\textbf{Feature Merging.}
Given the extracted 3D features $F^{\mathrm{3D}}$ and spherical projection features $F^{\mathrm{sph}}$, we first pass them through an $\ell_2$ normalization layer such that all features are constrained on a sphere. This is a common way to process multi-modal data \cite{clip}. We then formulate a fusion graph with complete edge connections, in which each node contains features from either $F^{\mathrm{3D}}$ or $F^{\mathrm{sph}}$. In addition, to enable features to interact directly with the global representation, we add two extra node features that are processed by the global average pooling module. The fusion graph node features are collected in the following set:
\begin{align}\label{eq:merge}
F = \set*{F^{\mathrm{3D}}, F^{\mathrm{sph}}, \mathrm{Pooling}(F^{\mathrm{3D}}), \mathrm{Pooling}(F^{\mathrm{sph}})}.
\end{align}

\textbf{Space-specific Interaction.}
We embed the fusion features $F$ into the Euclidean and hyperbolic spaces as
$F^{\mathbf{H}} = \mathrm{exp}^c_x(F)$ (where the $\exp$ operator is applied node-wise) and $F^{\mathbf{E}} = F$, respectively, to perform feature interaction using GA layers. Specifically, in the same way as \cref{eq:fc}, we obtain the $k$-th head FC features for $F^{\mathbf{H}}$ or $F^{\mathbf{E}}$, denoted as $F_k^{\mathrm{FC}}$.
We additionally leverage a learnable matrix $M$ regarded as a feature relationship metric such that the attention weights are computed as: 
\begin{align}\label{AMk}
A_{k}^{M}= \mathrm{Softmax}\parens*{ F_k^{\mathrm{FC}} M {F_k^{\mathrm{FC}}}\T }.
\end{align}
In Riemannian geometry, a Riemannian metric on a smooth manifold is a smooth symmetric covariant 2-tensor field that is positive definite at each point. The learnable matrix $M$ can be viewed as a more general extension of the Riemannian metric, where we do not impose any constraint on it, leaving it to update freely. The effectiveness of this design can be seen in \cref{tab:metric}, where the free metric surpasses other counterparts.

The learned feature embeddings from the Euclidean and hyperbolic spaces are then passed into two different GA layers and finally fused together using element-wise adding:
\begin{align}
F^{\mathrm{space}} = w^{\mathbf{E}} F^{\mathbf{E}} + w^{\mathbf{H}} F^{\mathbf{H}},
\end{align}
where $w^{\mathbf{E}}$ and $w^{\mathbf{H}}$ denote learnable weights for the Euclidean and hyperbolic embeddings $F^{\mathbf{E}}$ and $F^{\mathbf{H}}$, respectively. Each node feature has thus aggregated information from both Euclidean and hyperbolic spaces, which can be viewed as an adaptive combination of linearity and non-linearity that can contribute to a more effective feature representation.

\textbf{Modal-specific Interaction.}
We next decouple the merged features back to 3D features and projection features again, enabling them to turn around and learn information within their own modality. This is similar to the self-attention operation in the cross-self-attention pipeline. Specifically, for the 3D features, we pass them through a GA layer (with the learnable matrix) with preceding and succeeding MLP layers, while for the 2D projection features, we pass them through a basic ResNet block. After the modal-specific interaction, 3D features and projection features are merged together again using \cref{eq:merge} to reconstruct the fusion features.

\subsection{Pose Regression Head and Loss Function}
The task of LiDAR pose regression requires predicting a 6-DoF pose. However, since the translation and rotation elements do not scale compatibly, the regression converges in different basins. To deal with this problem, previous methods \cite{wang2021pointloc,memory-aware} consider the regression head with two parallel MLPs for translation and rotation regression, respectively. We thus use the same decoding head design as \cite{wang2021pointloc}, which consists of two MLP layers for translation and rotation regression as shown in \cref{fig:architecture}.  

During training, to provide sufficient supervision to the whole pipeline, we use not only the fusion features but also the 3D and projection features at lower levels. As shown in \cref{fig:architecture}, for the three features $F^{\mathrm{3D}}$, $F^{\mathrm{sph}}$, and $F$, we use three different regression heads $g^{\mathrm{3D}},g^{\mathrm{sph}},g$ respectively to predict their corresponding 6-DoF poses $(t^{\mathrm{3D}},r^{\mathrm{3D}})$, $(t^{\mathrm{sph}},r^{\mathrm{sph}})$, $(t,r)$.
Specifically, we first perform global average pooling and then regression to obtain the predicted poses, which can be described as follows:
\begin{align}\label{eq:regression}
\parens{t^{\mathrm{3D}},r^{\mathrm{3D}}}  & = \parens{g^{\mathrm{3D}}\circ\mathrm{Pooling})(F^{\mathrm{3D}}}, \\
\parens{t^{\mathrm{sph}},r^{\mathrm{sph}}} & = \parens{g^{\mathrm{sph}}\circ\mathrm{Pooling})(F^{\mathrm{sph}}}, \\
\parens{t,r} & = \parens{g\circ\mathrm{Pooling}}\parens{F},
\end{align}
where $\circ$ denotes the composition operation, and $\mathrm{Pooling}\parens*{}$ denotes the global average pooling operation. As for the rotation, we use the logarithmic format of the quaternion \cite{memory-aware,wang2021pointloc}. Denoting the translation and rotation targets as $t^*$ and $r^*$, the final loss function is computed as:
\begin{align}
&\ml{\mathcal{L}=(\norm{t^{\mathrm{3D}}-t^*}  + \norm{t^{\mathrm{sph}}-t^*} + \norm{t-t^*} ) e^{-\lambda} + \lambda \\
+ (\norm{r^{\mathrm{3D}}-r^*} + \norm{r^{\mathrm{sph}}-r^*}  + \norm{r-r^*} ) e^{-\gamma} + \gamma}
\end{align}
where $\lambda$ and $\gamma$ are learnable parameters. During inference, the pose $\parens*{t,r}$ predicted by the fusion features $F$ is treated as the final prediction.


\section{Experiments}
In this section, we first evaluate our proposed model on datasets collected from outdoors and indoors. We next present ablation studies to demonstrate the effectiveness of our model design.

\begin{table*}[t]\footnotesize
\centering
\begin{tabular}{ c | l | c  c   c  c     } 
\toprule
&  Model  &  Full-6  &  Full-7  &  Full-8  & Full-9 \\
\midrule
\textit{LiDAR Retrieval} 
&  PointNetVLAD\cite{uy2018pointnetvlad}  &  28.48 / 5.19  & 17.62 / 3.95  &  23.59 / 5.87  &  13.71 / 2.57  \\
\midrule
\textit{LiDAR Odometry} 
&  DCP\cite{dcp}  &  18.45 / 2.08   &  14.84 / 2.17   & 16.39 / 2.26   &  13.60 / 1.86   \\
\midrule
\multirow{4}{*}{\textit{Image-based PR}} 
&   PoseLSTM\cite{pose-lstm} &  26.36 / 6.54  &  74.00 / 9.85  &  128.25 / 18.59  &  19.12 / 3.05  \\
&   MapNet\cite{henriques2018mapnet}  &  48.21 / 6.06  &  61.01 / 5.85   &   75.35 / 9.67   &  44.34 / 4.54  \\
&   AD-MapNet\cite{ad-posenet}  &  18.43 / 3.28  &  19.18 / 3.95  &  66.21 / 9.42  &  15.10 / 1.82  \\
&   AtLoc+\cite{wang2020atloc}  &  17.92 / 4.73  &   34.03 / 4.01   &  71.51 / 9.91  &   10.53 / 1.97  \\
&   MS-Transformer\cite{ms-transformer}  &  11.69 / 5.66  &  65.38 / 9.01  &  88.63 / 19.80  & 7.62 / 2.53   \\
\midrule
\multirow{6}{*}{\textit{LiDAR-based PR}} 
&   PointLoc\cite{wang2021pointloc}  &  13.81 / \underline{1.53}  &  9.81 / \underline{1.27}  &  11.51 / \underline{1.34}  &  9.51 / \underline{1.07}  \\
&   PosePN\cite{memory-aware}  &  16.32 / 2.43   &   14.32 / 3.06   &  13.48 / 2.60  &   9.14 / 1.78  \\
&   PosePN++\cite{memory-aware}  &  10.64 / 1.78  &  9.59 / 1.92  &   \underline{9.01} / 1.51    &  8.44 / 1.71  \\
&   PoseSOE\cite{memory-aware}  &  \underline{8.81} / 2.04  &  \underline{7.59} / 1.94  &  9.21 / 2.12  &  \underline{7.27} / 1.87  \\
&   PoseMinkLoc\cite{memory-aware}  & 11.20 / 2.62   &  14.69 / 2.90  &  12.35 / 2.46  &  10.06 / 2.15  \\
&   HypLiLoc (ours)  &  \textbf{6.00} / \textbf{1.31}  &  \textbf{6.88} / \textbf{1.09}  &  \textbf{5.82} / \textbf{0.97}  &  \textbf{3.45} / \textbf{0.84}  \\
\bottomrule
\end{tabular}
\caption{Mean translation and rotation error (m/$^\circ$) on the Oxford Radar dataset. The best and the second-best results in each metric are highlighted in \textbf{bold} and \underline{underlined}, respectively. PR stands for pose regression. HypLiLoc achieves the best performance in all metrics.}
\label{tab:oxford}
\end{table*}

\begin{table*}[t]\footnotesize
\centering
\begin{tabular}{ c  | l | c c  c c  c c   } 
\toprule
&  Model  &  Seq-05  &  Seq-06  &  Seq-07  & Seq-14 \\
\midrule
\multirow{4}{*}{\textit{Image-based PR}} 
&   PoseLSTM\cite{pose-lstm}  &  0.16 / 4.23  &  0.18 / 5.28  &  0.24 / 7.05  &  0.13 / 4.81  \\
&   MapNet\cite{henriques2018mapnet}   &  0.26 / 6.67  &  0.28 / 6.91  &  0.39 / 9.17  &  0.25 / 6.85\\
&   AD-MapNet\cite{ad-posenet}  &   0.17 / 3.33  &  0.21 / 3.37  &  0.24 / 4.38  &  0.14 / 4.12  \\
&   AtLoc+\cite{wang2020atloc}  &  0.18 / 4.32  &  0.24 / 5.14  &  0.26 / 6.04  &  0.16 / 4.61  \\
&   MS-Transformer\cite{ms-transformer}  &  0.16 / 3.98  &  0.15 / 3.56  &  0.18 / 5.32  &  0.13 / 4.83  \\
\midrule
\multirow{6}{*}{\textit{LiDAR-based PR}} 
&   PointLoc\cite{wang2021pointloc}  & \underline{0.12} / 3.00   &  0.10 / 2.97   &    \textbf{0.13} / 3.47 &  0.11 / 2.84  \\
&   PosePN\cite{memory-aware}   &  \underline{0.12} / 4.38  & \underline{0.09} / 3.16  &  0.17 / 3.94  &  \textbf{0.08} / 3.27  \\
&   PosePN++\cite{memory-aware}  &  0.15 / 3.12  &  0.10 / 3.31  &  \underline{0.15} / \underline{2.92}   &  0.10 / \underline{2.80}   \\
&   PoseSOE\cite{memory-aware}  &  0.14 / 3.15  &  0.11 / \underline{2.90}  &  \underline{0.15} / 3.06  & 0.11 / 3.20 \\
&   PoseMinkLoc\cite{memory-aware}  &  0.16 / 5.17  & 0.11 / 3.74  &  0.21 / 5.74  &  0.12 / 3.64  \\
&   HypLiLoc (ours)   &  \textbf{0.09} / \textbf{2.52}  &  \textbf{0.08} / \textbf{2.58}  &  \textbf{0.13} / \textbf{2.55}  &  \underline{0.09} / \textbf{2.34}  \\
\bottomrule
\end{tabular}
\caption{Median translation and rotation error (m/$^\circ$) on the vReLoc dataset. The best and the second-best results in each metric are highlighted in \textbf{bold} and \underline{underlined}, respectively. PR stands for pose regression. HypLiLoc achieves the best performance in 7 out of 8 metrics.}
\label{tab:vreloc}
\end{table*}

\subsection{Implementation Details}
We use ResNet34 \cite{resnet}  pre-trained on ImageNet \cite{deng2009imagenet} as the backbone for projection features extraction. We use a batch size of $32$. The number of attention heads is set as $8$. Following \cite{hyp_img}, we set the base point $x$ as $0$ for hyperbolic embedding. We set $L^{\mathrm{3D}}=2$ and $L=2$. The Adam \cite{kingma2014adam} optimizer with the initial learning rate $1\times10^{-3}$ and weight decay $5\times10^{-4}$ is used for training. We train our network for $150$ epochs. All the experiments are conducted on either an NVIDIA RTX 3090 GPU or an NVIDIA RTX A5000 GPU.

\subsection{Datasets}
\textbf{Oxford Radar} is a large-scale outdoor autonomous driving dataset \cite{oxfordradar}. It provides data from multi-modal sensors, including LiDARs, cameras, Radars, and GPS, but in our experiments, we use only LiDAR information. It contains sensor data in the time span of $1$ year and a length span of $1000$ km. In addition, it covers various seasons and weather conditions, which thus allows a comprehensive evaluation of the models. Following \cite{wang2021pointloc,memory-aware}, we use the same benchmark data split setting, and we also report the mean translation rotation error.

\textbf{vReLoc} is an indoor robot dataset \cite{wang2021pointloc}. It consists of data from LiDARs, cameras, depth cameras, and motion trackers. In our experiments, we use only LiDAR information. It contains both static and dynamic scenarios with people walking around. Following \cite{wang2021pointloc,memory-aware}, we use the same benchmark data split setting, and we also report the median translation and rotation error.

\subsection{Main Results}
We first compare HypLiLoc with other baselines on the Oxford Radar dataset. From \cref{tab:oxford}, we observe that HypLiLoc achieves SOTA performance in all metrics. Especially on the route Full-9, HypLiLoc obtains $3.45$ m mean translation error in the city-wise relocalization task compared to the second best performer PoseSOE with $7.27$ m, which demonstrates the effectiveness of HypLiLoc. In addition, compared with camera pose regression approaches that take images as inputs, LiDAR-based ones are generally more accurate. This verifies that point clouds generated by LiDARs are a more effective data modality for the relocalization task. \cref{tab:oxford} also indicates that for large-scale pose estimation, LiDAR pose regression approaches surpass both retrieval-based and odometry-based ones, and thus this approach is promising for many applications. 

Note that pose regression approaches can be integrated into SLAM systems \cite{akilan2020multimodality,ali2022bi,li2018ongoing} to achieve even better accuracy and to perform fast global pose estimating, especially in cases where a global navigation satellite system is not available (e.g., indoors and urban areas with dense skyscrapers).

We next test HypLiLoc on the indoor vReLoc dataset in \cref{tab:vreloc}, where it achieves SOTA performance in $7$ out of $8$ metrics and shows strong competitiveness. We note in the indoor environment, the LiDAR-based approaches also generally outperform image-based ones.

\subsection{Analysis}\label{subsec:analysis}
\textbf{Ablation Study.}
We provide insights into our design choices for HypLiLoc by ablating each module. From \cref{tab:ablation_study_modules}, every module in our design contributes to the final improved estimation accuracy. Making use of information from the projected point cloud image and hyperbolic-Euclidean feature fusion strategy both contribute to more accurate pose regression outputs. 

\begin{table}\footnotesize
\centering
\begin{tabular}{l  c  c } 
\toprule
Method &  Mean Error (m/$^\circ$) on Full-8 \\
\midrule
base model &  9.78 / 1.99  \\
+ global graph attention  &    8.91 / 1.74  \\
+ spherical-projection backbone  &  7.26 / 1.36   \\
+ feature fusion block  &  6.19 / 1.13    \\
+ learnable metric (full model)  &   \textbf{5.82} / \textbf{0.97}   \\
\midrule
full model w/o hyperbolic branch  &   6.57 / 1.19   \\
full model w/o Euclidean branch  &   6.24 / 1.16   \\
\bottomrule
\end{tabular}
\caption{Ablation study for different modules on Full-8 route of the Oxford Radar dataset.}
\label{tab:ablation_study_modules}
\end{table}

\textbf{Different Projection Strategies.}\label{subsubsec:projection}
We next compare different modality strategies. As shown in \cref{tab:projection}, we first test the performance using the single modality input, including the 3D point cloud, the spherical projection, and the BEV projection. Among them, the 3D and the spherical projection show similar performances, while the BEV performance is worse. This verifies our insight that the BEV projection is not a bijective mapping, and thus less information is retained. 

When feeding two modalities, the combination \textit{3D + spherical} surpasses the other two counterparts, which is our final model choice for HypLiLoc. When we further add the BEV input, the performance drops instead.

\begin{table}\scriptsize
\centering
\begin{tabular}{c  |  c  c  c   c   c } 
\toprule
\#Modalities &  3D  & Sph.  &  BEV    & Mean Error (m/$^\circ$) on Full-8\\
\midrule
\multirow{3}{*}{1} & \checkmark  &    &    &  8.91 / 1.74    \\
& &  \checkmark   &    &  8.94 / 2.18\\
& &    &  \checkmark   &  9.46 / 2.33\\
\midrule
\multirow{3}{*}{2} & \checkmark  &  \checkmark   &     & \textbf{5.82} / \textbf{0.97}  \\
& \checkmark  &     &  \checkmark  &  9.44 / 1.80\\
&  &  \checkmark   &  \checkmark  &  6.77 / 1.02\\
\midrule
\multirow{1}{*}{3} & \checkmark  &  \checkmark   &  \checkmark  &  6.32 / 1.01\\
\bottomrule
\end{tabular}
\caption{Comparison of different projection methods on Full-8 route of the Oxford Radar dataset. For the single modality, we do not use the feature fusion block.
}
\label{tab:projection}
\end{table}


\textbf{Learnable Matrix $M$ Design.}
We test the performance by applying different constraints on the learnable matrix $M$ in \cref{AMk}. The Riemannian metric, which formulates a positive definite and symmetric matrix, has the strictest constraints. However, as shown in \cref{tab:metric},  the Riemannian metric does not provide performance improvements compared with the setting without any metric. If we only impose either the positive definite constraint or the symmetric constraint, the performance improves.
Furthermore, if we exclude all constraints and enable the metric to evolve freely, we can achieve optimal performance.

\begin{table}\footnotesize
\centering
\begin{tabular}{l  c  c } 
\toprule
Method &  Mean Error (m/$^\circ$) on Full-8 \\
\midrule
w/o $M$ &  6.19 / 1.13 \\
Riemannian &  6.34 / 1.28 \\
positive definite  &   6.18 / 1.13  \\
symmetric  &  6.02 / 1.24  \\
no constraint  &  \textbf{5.82} / \textbf{0.97} \\
\bottomrule
\end{tabular}
\caption{Comparison of different constraints on Full-8 route of the Oxford Radar dataset.}
\label{tab:metric}
\end{table}

\textbf{Computational Time and Storage.}
We compare LiDAR-based models that belong to different relocalization pipelines. As observed in \cref{tab:total_memory}, regression-based models can operate at least $2$ times as fast as both the retrieval-based and odometry-based approaches. For the runtime memory, regression-based models need only $1/3$ that of retrieval and odometry methods (less than $7$ GB). HypLiLoc can perform inference at a speed of $48$ FPS, which is $4$ times as fast as than the retrieval model and over $2$ times as fast as PointLoc.  applications.

\begin{table}\scriptsize
\centering
\begin{tabular}{c l  c  c  c} 
\toprule
& \multirow{2}{*}{Model}  &  Runtime  & Runtime \\
& &   Speed & Total Memory\\
\midrule
\textit{Retrieval} &  PointNetVLAD\cite{uy2018pointnetvlad}  &  11FPS & 26GB\\
\midrule
\textit{Odometry}&  DCP\cite{dcp}  & 10FPS  & 22GB\\
\midrule
\multirow{2}{*}{\textit{Regression}}&  PointLoc\cite{wang2021pointloc}  & 22FPS  & 7GB\\
&  HypLiLoc (ours)  &  48FPS  & 6GB\\
\bottomrule
\end{tabular}
\caption{Comparison of the runtime speed and the runtime total memory of different models. }
\label{tab:total_memory}
\end{table}


\begin{figure}[!htb]
\begin{center}
\includegraphics[width=0.3\textwidth]{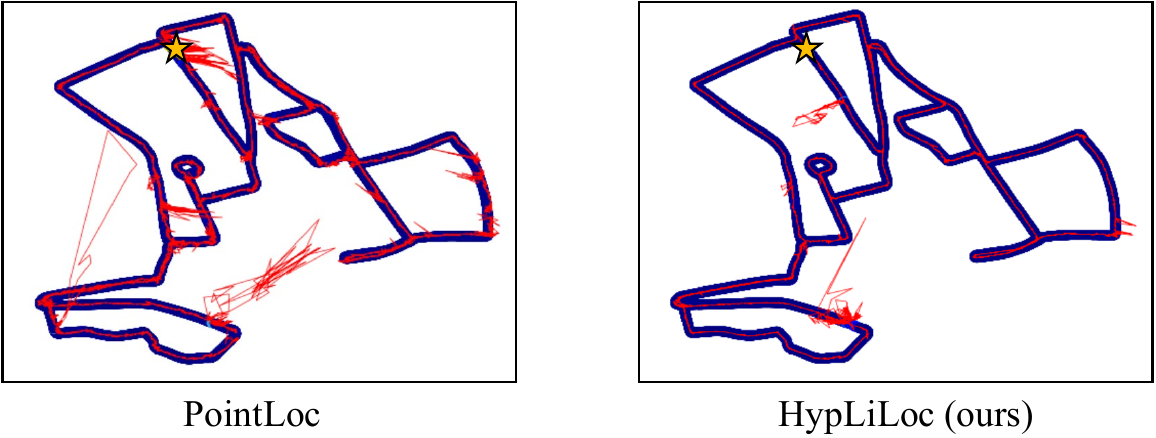}
\end{center}
\caption{Trajectory visualization on the Oxford Radar dataset. The ground truth trajectories are shown in bold blue lines, and the estimated trajectories are shown in thin red lines. }
\label{fig:trajectory}
\end{figure}

\textbf{Visualization.}
We visualize a typical output pose trajectory of HypLiLoc and PointLoc in \cref{fig:trajectory}. HypLiLoc outputs a smoother and more accurate pose trajectory compared with PointLoc.

\section{Limitations}\label{sec:limitations}
Although we have tested the proposed model in a city-wise dataset, verification of HypLiLoc's performance in challenging scenarios (e.g. with noise perturbations and adversarial attacks) is necessary for practical implementations. 

\section{Conclusion}
In this work, we propose HypLiLoc, a novel network for LiDAR-based pose regression. It achieves effective feature extraction with global graph attention, hyperbolic-Euclidean interaction, and modal-specific learning. It achieves SOTA performance in both outdoor and indoor datasets.

\section{Acknowledgment}
This research is supported by A*STAR under its RIE2020 Advanced Manufacturing and Engineering (AME) Industry Alignment Fund – Pre Positioning (IAF-PP) (Grant No. A19D6a0053) and the National Research Foundation, Singapore and Infocomm Media Development Authority under its Future Communications Research and Development Programme. The computational work for this article was partially performed on resources of the National Supercomputing Centre, Singapore (\url{https://www.nscc.sg}).


{\small
\bibliographystyle{ieee_fullname}
\bibliography{egbib}
}

\newpage

\begin{center}
  {
  \large
  \lineskip .5em
  \begin{tabular}[t]{c}
    \textbf{Supplement}
  \end{tabular}
  \par
  }
  \vskip .5em
  \vspace*{12pt}
\end{center}


\begin{figure}[!htb]
\begin{center}
\includegraphics[width=0.45\textwidth]{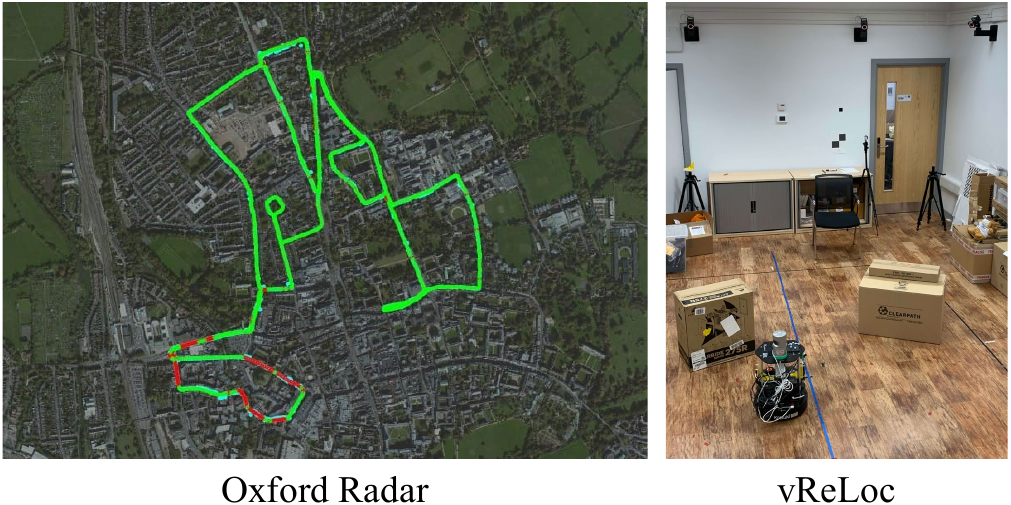}
\end{center}
\caption{Overview of the two datasets.}
\label{sfig:overview}
\end{figure}

\begin{figure}[!htb]
\begin{center}
\includegraphics[width=0.45\textwidth]{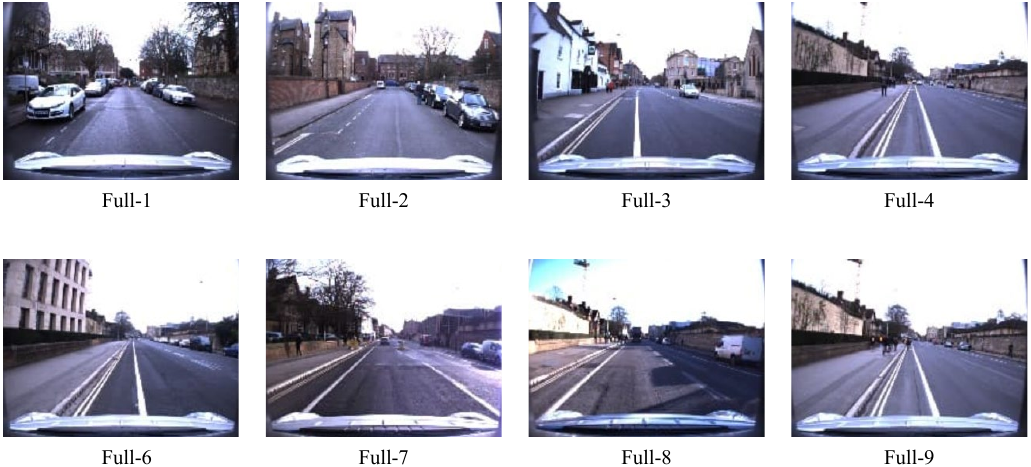}
\end{center}
\caption{Visualization of the Oxford Radar dataset. The RGB camera images are for visualization only, and we do not use them in our pipeline.}
\label{sfig:oxford_viz}
\end{figure}

\begin{figure*}[!htb]
\begin{center}
\includegraphics[width=0.95\textwidth]{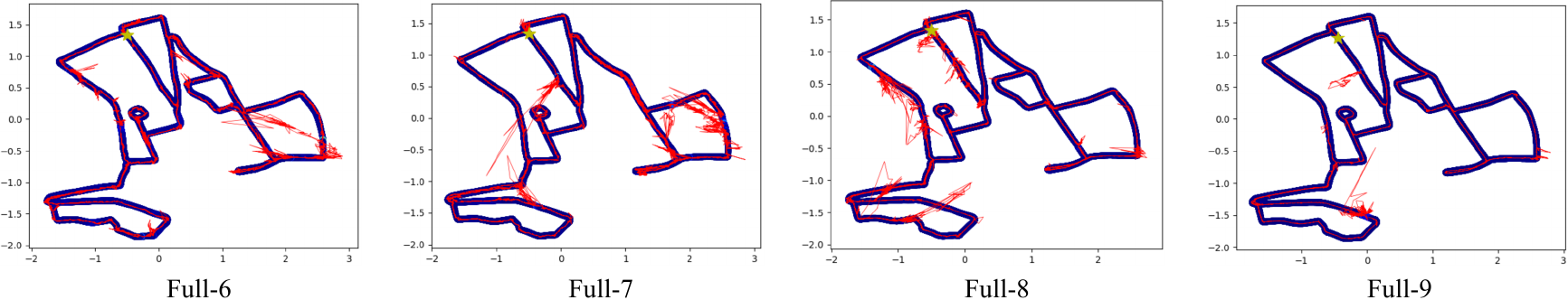}
\end{center}
\caption{More trajectory visualization on the Oxford Radar dataset. The ground truth trajectories are shown in bold blue lines, and the estimated trajectories are shown in thin red lines. }
\label{sfig:oxford_traj}
\end{figure*}

\section{Supplement Video and Code}
We welcome readers to watch our supplemental anonymous video that shows the runtime performance on the Oxford Radar dataset:
\url{https://www.youtube.com/watch?v=qplZMOZG-7k}

We also welcome readers to run our code at:
\url{https://github.com/sijieaaa/HypLiLoc}

\section{More About Pose Regression}
Visual relocalization refers to the process of determining visual sensor poses from known scene representations such as images, point clouds, key points, features, pose maps, and neural networks.

The LiDAR-based pose regression is a type of relocalization pipeline, where the known scene is represented by neural networks in an implicit way, which is different from previous solutions. Given LiDAR point clouds as inputs, the neural network regresses the corresponding poses directly. This formulation is similar to (but operates reversely) the current popular Neural Radiance Fields (NeRFs) that take poses as inputs and outputs the corresponding sensor data.
Therefore, the pose regression network can be viewed as another type of neural representation.

We compare typical localization pipelines in \cref{stab:different_methods}. Structure-based methods achieve the highest global/local pose accuracy, but they suffer from the lowest speed. These types of methods are usually used for offline applications where high-speed inference is not necessary and sufficient computing resources are provided. Visual odometry methods estimate relative poses between frames and serve as a module in the complete SLAM system. The SLAM system provides accurate local pose estimations, but global pose estimations depend on additional information such as loop closure. The retrieval-based methods predict the pose by exhaustively searching the top-matched representations in the database, which is the main cause of high memory consumption and low inference speed. By contrast, pose regression models implicitly represent the scene using neural networks and do not require the database during inference.

\begin{table*}[!htb]\footnotesize
\centering
\begin{tabular}{l   c  c  c  c  c} 
\toprule
Pipeline  & 3D Model & Inference Database  & Global Acc. & Local Acc. & Speed \\
\midrule
Structure-based keypoint matching  &  need  &  need &  high  &  high  & slow  \\
Visual odometry &  - &  - &  low  &  high  &  fast \\
SLAM & increasing  &  increasing  &  low   &  high  &  medium   \\
Retrieval & -  &   need &   medium   &  medium &  slow \\
Pose regression & -  &  - & medium  &  medium   &  fast \\
\bottomrule
\end{tabular}
\caption{Comparison of different localization pipelines.}
\label{stab:different_methods}
\end{table*}

\section{Performance After Outlier Filtering}
Retrieval-based models require a pre-built database to store candidate scene representations with corresponding poses. Pose regression models do not rely on any database but may suffer from extreme outliers that are far from roads because there is no map or road trajectory information provided. In contrast, the retrieval-based models can only output the locations that are restricted to the trajectory.

This inspires us to explore the possibility of including trajectory information for our regression-based model to further improve its performance. We exclude outlier poses that are far from the database poses over some threshold distances. 
More specifically, if the regression-based model outputs some pose that is far from all the poses that have been recorded in the trajectory over a threshold distance, we take it as an outlier and discard this output. (Note this technique is only introduced in the supplement and is not used in the main paper.)

In \cref{stab:outlier_full8}, outlier filtering brings $1.25$m/$0.31^{\circ}$ mean error improvements and excludes $27.6\%$ pose estimations. In \cref{stab:outlier_full9}, this strategy even supports our network to achieve less than $3$m translation error with $16.1\%$ pose estimations dropped out, which is a promising result in the city-wise relocalization task. 

In real applications, the exclusion of outliers can be augmented with other techniques like the wheel or LiDAR odometry modules to remedy the dropped poses.

\begin{table}[!htb]\footnotesize
\centering
\begin{tabular}{c  c  c  c  c  c} 
\toprule
Outlier Thd. (m)  & Mean Error (m/$^\circ$) & Remaining Poses (\%) \\
\midrule
None  &  5.82 / 0.97  & 100.0  \\
25  &  5.37 / 0.88  &   99.5\\
10  &  5.10 / 0.82  &   98.6    \\
7   &  5.03 / 0.79  &   98.1  \\
5   &  4.96 / 0.77  &   97.4     \\
3   &  4.85 / 0.75  &   95.2    \\
1   &  \textbf{4.57} / \textbf{0.66}  &   72.4    \\
Difference    &  (\textbf{-1.25} / \textbf{-0.31})  & (-27.6)    \\
\midrule
PointNetVLAD  &  23.59 / 5.87  & 100.0  \\
\bottomrule
\end{tabular}
\caption{Performance after filtering pose estimation outliers on Full-8 route of the Oxford Radar dataset.}
\label{stab:outlier_full8}
\end{table}

\begin{table}[!htb]\footnotesize
\centering
\begin{tabular}{c  c  c  c  c  c} 
\toprule
Outlier Thd. (m)  & Mean Error (m/$^\circ$) & Remaining Poses (\%) \\
\midrule
None  &  3.45 / 0.84  & 100.0  \\
25  &  3.27 / 0.74  &  99.5 \\
10  &  3.18 / 0.71  &  99.2      \\
7   &  3.15 / 0.70  &  99.0    \\
5   &  3.11 / 0.69  &  98.8   \\
3   &  3.05 / 0.68  &  97.7      \\
1   &  \textbf{2.90} / \textbf{0.64}  &  83.9      \\
Difference    &  (\textbf{-0.55} / \textbf{-0.20})  & (-16.1)    \\
\midrule
PointNetVLAD  &  13.71 / 2.57  & 100.0  \\
\bottomrule
\end{tabular}
\caption{Performance after filtering pose estimation outliers on Full-9 route of the Oxford Radar dataset.}
\label{stab:outlier_full9}
\end{table}

\section{More Trajectory Visualization}
We visualize more of the output trajectories from different routes on the Oxford Radar dataset as shown in \cref{sfig:oxford_traj}.

\section{Dataset Details}
The datasets we used in our experiments include the Oxford Radar dataset and the vReLoc dataset as shown in \cref{sfig:overview}. For the Oxford Radar dataset, we also visualize the environmental conditions on different routes in \cref{sfig:oxford_viz}. Note that the RGB camera images are for visualization only, and we do not use them in our network. Both of the datasets are available online at:
\begin{itemize}
\item \url{https://oxford-robotics-institute.github.io/radar-robotcar-dataset/}
\item \url{https://github.com/loveoxford/vReLoc}
\end{itemize}

For each dataset, we list the corresponding data split as shown in \cref{stab:oxford config} and \cref{stab:vreloc config}.

\begin{table}[!htb]\footnotesize
\centering
\begin{tabular}{c  c  c  c  c  c} 
\toprule
Scene & Date/Time & Tag & Training & Test \\
\midrule
Full-1 &  2019-01-11-14-02-26 & sun      &    \checkmark &  \\
Full-2 &  2019-01-14-12-05-52 & overcast &  \checkmark &  \\
Full-3 &  2019-01-14-14-48-55 & overcast &  \checkmark &  \\
Full-4 &  2019-01-18-15-20-12 & overcast &  \checkmark &  \\
\midrule
Full-6 &  2019-01-10-11-46-21 & rain &   &  \checkmark \\
Full-7 &  2019-01-15-13-06-37 & overcast &  & \checkmark  \\
Full-8 &  2019-01-17-14-03-00 & sun &   &  \checkmark \\
Full-9 &  2019-01-18-14-14-42 & overcast &  & \checkmark  \\
\bottomrule
\end{tabular}
\caption{Dataset details on the Oxford Radar dataset.}
\label{stab:oxford config}
\end{table}

\begin{table}[!htb]\footnotesize
\centering
\begin{tabular}{c  c  c  c  c  c} 
\toprule
Scene & Tag & Training & Test \\
\midrule
Seq-03 &  static  &  \checkmark &  \\
Seq-12 &  walking &  \checkmark &  \\
Seq-15 &  walking &  \checkmark &  \\
Seq-16 &  walking &  \checkmark &  \\
\midrule
Seq-05 &  static &   &  \checkmark \\
Seq-06 &  static &   &  \checkmark \\
Seq-07 &  static &   &  \checkmark \\
Seq-14 &  walking &  &  \checkmark \\
\bottomrule
\end{tabular}
\caption{Dataset details on the vReLoc dataset.}
\label{stab:vreloc config}
\end{table}

\section{Baseline Models}
The baseline models in our comparison include: PointNetVLAD, DCP, PoseLSTM, MapNet, AD-MapNet, AtLoc+, MS-Transformer, PointLoc, PosePN, PosePN+, PoseSOE, and PoseMinkLoc.

\section{Codebase}
Our codes are developed based on the following repositories:
\begin{itemize}
\item \url{https://github.com/ori-mrg/robotcar-dataset-sdk},
\item \url{https://github.com/BingCS/AtLoc},
\item \url{https://github.com/htdt/hyp_metric}.
\end{itemize}

\end{document}